\documentclass[10pt]{article}
\usepackage[letterpaper]{geometry}
\usepackage{hicss}
\usepackage{times}
\usepackage[none]{hyphenat}
\usepackage{url}
\usepackage{latexsym}
\usepackage{minted}
\usepackage{indentfirst}
\usepackage{graphicx}
\graphicspath{{images/}}
\usepackage[
    style=apa,
  ]{biblatex}
\usepackage[style=apa]{biblatex} 

\usepackage{graphicx}
\usepackage[hidelinks]{hyperref}
\newcommand{\mypar}[1]{\smallskip\noindent\textbf{#1.}}
\newcommand{\mypartwo}[1]{\vspace{0.5pt}\noindent\textit{#1.}}

\usepackage{amsfonts}
\usepackage{booktabs}
\usepackage{enumitem}
\usepackage{amsmath}
\usepackage{bm}
\usepackage{subcaption}
\usepackage{tabularx}
\usepackage{multirow}
\usepackage{multicol}
\usepackage{arydshln}
\usepackage{wrapfig}

\usepackage{xcolor}

\usepackage{booktabs}
\usepackage{arydshln}

\usepackage{tikz}
\usetikzlibrary{positioning,arrows.meta}

\usepackage{marginnote}

\title{On the Potential of Multi-Task Learning in Predictive Process Monitoring}

\author{
\begin{tabular}{cc}
\begin{tabular}{c}
Lukas Kirchdorfer\textsuperscript{*} \\
SAP Signavio, University of Mannheim \\
\underline{lukas.kirchdorfer@sap.com}
\end{tabular}
&
\begin{tabular}{c}
Keyvan Amiri Elyasi\textsuperscript{*} \\
University of Mannheim \\
\underline{keyvan.amiri@uni-mannheim.de}
\end{tabular}
\\[3em]
\multicolumn{2}{c}{
\begin{tabular}{c}
Heiner Stuckenschmidt \\
University of Mannheim \\
\underline{heiner.stuckenschmidt@uni-mannheim.de}
\end{tabular}
}
\\[1em]
\multicolumn{2}{c}{
\small\textsuperscript{*}Equal contribution
}
\end{tabular}
}

\begin{document}
\maketitle
\begin{abstract}
Predictive Process Monitoring (PPM) forecasts how ongoing organizational processes unfold, enabling information systems to move beyond execution support toward proactive analysis and monitoring. Although deep learning has improved prediction accuracy in PPM, most approaches follow a single-task learning (STL) setup, training a separate model per task. This increases maintenance effort and overlooks potential synergies.
Multi-task learning (MTL), which jointly learns multiple prediction targets in one model, offers a promising alternative, yet its effectiveness in PPM remains underexplored. It remains unclear whether and under which settings MTL improves upon STL, which prediction tasks benefit most from joint learning, which task combinations are particularly synergistic, and if and how tasks should be balanced.
To fill this gap, we present the first comprehensive empirical study of MTL for PPM, evaluating a variety of task combinations, neural architectures, and optimization methods.
Overall, our results position MTL as a strong paradigm for PPM: we see substantial improvements in next-activity prediction and inherent mitigation of class imbalance using MTL, while task balancing is especially critical under low-capacity models.
\end{abstract}

\subsubsection*{Keywords:}

Predictive process monitoring, Multi-task learning, Deep learning, Process mining.

\section{Introduction}
\label{sec:intro}
Information systems support the execution of organizational processes across diverse domains, and generate extensive streams of event data that record sequences of human and machine activities over time~\parencite{Aalst16}. Anticipating how these processes will unfold, is crucial for optimizing resource allocation, preventing failures, and enabling timely operational decisions. Predictive process monitoring (PPM) exploits historical execution data to forecast key aspects of ongoing cases, such as the next activity to occur, the timing of the next event, the remaining cycle time, or the eventual outcome \parencite{di2022predictive}.
Recent progress in PPM has been driven by deep learning models, typically developed under a single-task learning (STL) paradigm, focusing on a single prediction target such as estimating the remaining time of an ongoing case.

Despite their strong predictive performance, STL models 
face two key limitations. First, they are computationally expensive \parencite{ChenBLR18}, hindering deployment in real-world industrial settings. For instance, process mining vendors such as SAP Signavio, Celonis, or Apromore would need to train and maintain separate models for every task, process, and customer, resulting in increased operational overhead and limited scalability. Second, STL models treat prediction tasks in isolation and overlook cross-task dependencies \parencite{kendall2018multi,ChenBLR18}. For example, in a hospital process, knowing the next activity can support remaining-time prediction and vice versa. Ignoring such synergies prevents STL models from fully leveraging the structure of process data.

To overcome these limitations, multi-task learning (MTL) offers a promising alternative, enabling a single model to jointly learn multiple prediction targets \parencite{caruana1997multitask}. 
This paradigm has proven highly effective in domains such as computer vision \parencite{kendall2018multi}. 
However, in the context of PPM, the potential of MTL remains largely underexplored. 
The few studies that adopt an MTL setup \parencite{TaxVRD17,Camargo2019,wuyts2024sutran,hennig2025leveraging} provide little evidence on whether and when MTL improves predictive performance. It remains unclear how individual tasks behave when trained jointly, and which combinations of PPM tasks are actually complementary. Moreover, none of the existing works employ task-balancing techniques, which are often a critical success factor for MTL \parencite{kendall2018multi}. Instead, they rely on unweighted sums of task losses, allowing dominant tasks to suppress others and potentially degrading overall performance.


The main contribution of this paper is to address these open questions through the first systematic and large-scale empirical evaluation of MTL for PPM. Our study provides actionable insights into when and how MTL is beneficial. To this end, we conduct an extensive benchmark across nine real-world event logs from diverse domains, three neural network architectures, all pairwise and joint combinations of three common predictive tasks 
(next activity, next time and remaining time prediction),
and thirteen multi-task optimization (MTO) methods.
Our study is guided by the following research questions:

\begin{itemize}[noitemsep,topsep=0pt,leftmargin=*]
\item \textbf{RQ1}: Does MTL outperform STL, and which neural network architectures benefit most from the multi-task setup?
\item \textbf{RQ2}: Which prediction targets are most and least affected by joint learning?
\item \textbf{RQ3}: How does performance vary across different task combinations?
\item \textbf{RQ4}: Which MTO methods yield the best results?
\end{itemize}


\section{Background and Related Work}
\label{sec:background}
This section presents the background on MTL and reviews related work on applications of MTL in PPM. 

\mypar{Multi-Task Learning}
MTL refers to the paradigm of jointly training a model on multiple related prediction tasks, with the aim of improving generalization through shared representations and inductive transfer \parencite{caruana1997multitask}. 
Research on MTL can broadly be categorized into three main areas. The first line of work focuses on \emph{network architectures}, particularly on how features should be shared across tasks to maximize performance \parencite{liu2019end}.
The second area investigates \emph{task affinities}, which examine a grouping of tasks that should be learned together to benefit from the joint training \parencite{standley2020tasks}.
The third area, \emph{multi-task optimization} (MTO), focuses on balancing tasks during training to mitigate negative transfer, where certain tasks dominate training and degrade performance on others. This issue can arise from conflicting gradients, where task-specific gradients differ substantially in magnitude or direction. When such conflicts occur, updating the shared backbone using the average gradient can harm one or more tasks.

A plethora of MTO methods have been proposed to mitigate negative transfer. 
\textit{Loss weighting methods} aim to balance the training by assigning individual weights to the task-specific losses. These weights can be derived by either learning task uncertainties (UW) \parencite{kendall2018multi}, computing the rate of change of losses (DWA) \parencite{liu2019end}, random sampling (RLW) \parencite{RLW}, computing the (normalized) inverse of the losses as weights (UW-O and UW-SO) \parencite{KirchdorferEKSSK22}, or performing a grid search over different possible weight combinations (Scalarization) \parencite{xin2022current}. The geometric loss strategy (GLS) \parencite{chennupati2019multinet++} offers an alternative by combining the losses through their geometric mean.
\textit{Gradient-based methods} instead operate directly on task-wise gradients, either by rescaling them or by enforcing alignment before aggregation. Gradient-scaling approaches include Grad-Norm \parencite{ChenBLR18}, which normalizes gradients to balance task influence, and Nash-MTL \parencite{NavonSAMKCF22}, which frames MTL as a bargaining problem and derives corresponding scaling factors. Alignment-oriented techniques—such as GradDrop \parencite{ChenNHLKCA20}, PCGrad \parencite{YuK0LHF20}, and CAGrad \parencite{LiuLJSL21}—manipulate gradient vectors to reduce conflicts between tasks.
Finally, \textit{hybrid methods} such as IMTL \parencite{LiuLKXCYLZ21} combine loss weighting with gradient scaling to leverage the advantages of both families.


\begin{table}[t!]
\centering
\scriptsize
\setlength{\tabcolsep}{3pt}
\caption{Comparison of existing PPM approaches that apply MTL.
Arch.: L = LSTM, T = Transformer, C = CNN. 
STL indicates whether MTL is compared against its single-task baseline. 
Com. indicates whether all task combinations are evaluated. 
Analysis indicates whether the study analyzes when and why MTL is beneficial. 
}
\label{tab:mtl_ppm_related_work}
\begin{tabular}{lllcccc}
\toprule
 & Tasks & Arch. & STL & Com. & MTO & Analysis \\
\midrule
\textcite{TaxVRD17} & 2 & L & \checkmark &  & EW &  \\
\textcite{gherissi2022object} & 3 & L &  &  & EW &  \\
\textcite{Camargo2019} & 3 & L & \checkmark & $\sim$ & EW &  \\
\textcite{TaymouriREBV20} & 2 & L &  &  & EW &  \\
\textcite{hennig2025leveraging} & 3 & T &  &  & EW &  \\
\textcite{wang2024multi} & 3 & T &  &  & static &  \\
\textcite{wuyts2024sutran} & 3 & T & &  & EW &  \\
\textcite{WangHMLWY23} & 3 & T &  &  & static &  \\
\textcite{LazoN22} & 3 & T & \checkmark  & $\sim$ & EW &  \\
\textcite{GunnarssonBW23} & 3 & L &  &  & EW &  \\
\midrule
\textbf{Our work} & 3 & L/C/T & \checkmark & \checkmark & 13 & \checkmark \\
\bottomrule
\end{tabular}
\label{tab:rel}
\end{table}

\mypar{MTL in PPM}
MTL has already been applied in PPM approaches, but only in a limited number of studies, as summarized in Table~\ref{tab:rel}. Early work mainly relied on LSTM-based models that jointly predict the next activity and time, while more recent studies explore multi-task Transformer-based models.
Prior studies mainly focus on a single architecture and specific task combinations, typically rely on equal or static task weights, rarely compare against the STL baselines, and perform no further analysis of the MTL setup. In contrast, our work constitutes the first comprehensive benchmark of MTL in PPM by (i) analyzing in depth when and how MTL is beneficial, (ii) evaluating 13 MTO methods rather than only equal or static weighting, (iii) systematically investigating all combinations of three predictive tasks, and (iv) comparing MTL against STL across LSTMs, CNNs, and Transformers.
\section{Preliminaries}
\label{sec:preliminaries}
This section provides a formal introduction to the data, the PPM tasks, and the MTL problem statement.

\mypar{Event Log, Traces, Event Prefixes}
We assume that process execution data is recorded in an event log $\mathcal{E}$, defined as a finite set of traces. Each trace  $\sigma = \langle e_{1}, e_{2}, ..., e_{n}\rangle$ represents the ordered sequence of events for a single process instance (case), and each event is a tuple  $e = (c, a, t, r)$. Let $\pi_A$ and $\pi_T$ denote functions mapping the event to its activity and timestamp, such that $\pi_A(e) = a$ and $\pi_T(e) = t$. The remaining event attributes $c$ and $r$ correspond to the case identifier and the responsible resource, respectively. For a given trace $\sigma$, its event prefix of length $m \in [1, n-1]$ is defined as $\sigma^m=\langle e_{1}, \ldots, e_{m}\rangle$, representing the partial execution of the case up to the $m$-th event.

\mypar{PPM Tasks}
PPM is commonly formulated as either a classification or regression problem, typically beginning with a feature extraction step. In this step, event prefixes of varying lengths are generated from each completed trace to represent the process at different execution stages. Each event prefix is then encoded into a feature vector $\Gamma(\sigma^m) = \bm{x} \in \mathcal{X}$ and associated with one or more target values $y_k \in \mathcal{Y}$, each corresponding to a specific PPM task. Given an event prefix $\sigma^m$ of a trace $\sigma$, we consider the following three commonly studied PPM tasks, each defined as a predictive function over the feature vector: 

\begin{itemize}[noitemsep,topsep=0pt,leftmargin=*]
    \item \textbf{Next activity prediction} learns a function $\delta_a$ to predict the next activity: $\hat{y}_1 = \delta_{a}(\bm{x}) \approx \pi_A(e_{m+1})$.   

    \item \textbf{Next time prediction} learns a function $\delta_{t}$ to estimate the time interval to the next event: $\hat{y}_2 = \delta_{t}(\bm{x}) \approx \pi_T(e_{m+1}) - \pi_T(e_m)$.
    
    \item \textbf{Remaining time prediction} learns a function $\delta_{rt}$ to estimate the time interval between the current event and the last event of the trace: $\hat{y}_3 = \delta_{rt}(\bm{x}) \approx \pi_T(e_{n}) - \pi_T(e_m)$. 
 
\end{itemize}

\begin{figure}[t]
    \centering
    \resizebox{\linewidth}{!}{
    \begin{tikzpicture}[
        box/.style={
            draw,
            rounded corners,
            align=center,
            minimum height=0.8cm,
            minimum width=2.2cm,
            font=\small
        },
        arrow/.style={-Latex, thick},
        node distance=1.2cm
    ]

    \node[box, dashed] (input) {Event prefix\\ $\bm{x}$};
    \node[box, right=of input, ] (backbone) {Shared\\ backbone};

    \node[box, right=1.4cm of backbone, yshift=0.9cm] (nap) {NAP head};
    \node[box, right=1.4cm of backbone] (ntp) {NTP head};
    \node[box, right=1.4cm of backbone, yshift=-0.9cm] (rtp) {RTP head};

    \draw[arrow] (input) -- (backbone);
    \draw[arrow] (backbone.east) -- (nap.west);
    \draw[arrow] (backbone.east) -- (ntp.west);
    \draw[arrow] (backbone.east) -- (rtp.west);

    \end{tikzpicture}
    }
    \caption{Multi-task setup with shared backbone (LSTM/CNN/Transformer) and task-specific heads.}
    \label{fig:mtl_architecture}
\end{figure}
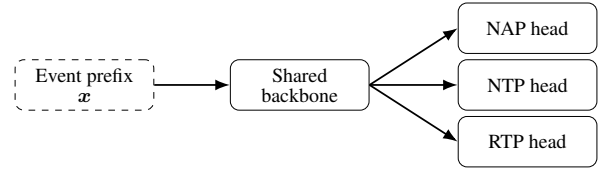

\mypar{MTL Problem Statement}
We consider supervised multi-task learning with a single input domain and a neural network composed of a shared backbone and task-specific heads, as illustrated in Figure~\ref{fig:mtl_architecture}. Given an input $\bm{x} \in \mathcal{X}$, the model returns one prediction $y_k$ per task $k \in \{1,\dots,K\}$. For our three PPM tasks, this corresponds to
$f_{\bm{\theta}}(\bm{x}) = [\delta_{a}(\bm{x}), \delta_{t}(\bm{x}), \delta_{rt}(\bm{x})]$, with $\bm{\theta}$ denoting the trainable parameters (shared and task-specific ones).
Each task has its own loss $\mathcal{L}_k(f_{\bm{\theta}}(\bm{x}), y_k)$ (e.g., cross-entropy for classification or mean squared error for regression). During training, we combine them with a weighted sum
    $\mathcal{L} = \sum_k \omega_k \mathcal{L}_k$,
where $\omega_k \ge 0$ are task weights (e.g., for EW: $\omega_k = 1$ ). 

\section{Experimental Setup}
\label{sec:setup}
This section outlines the experimental design used to evaluate the effectiveness of a multi-task setup for training predictive process monitoring models. The full implementation (developed in Python and PyTorch), along with supplementary material and additional results, is publicly available in our repository\footnote{\url{https://github.com/lukaskirchdorfer/MTL4PPM}}.

\mypar{Evaluation Data}
Our evaluation uses nine real-world event logs,\footnote{Datasets from \url{https://data.4tu.nl}; available in our repository} as described in \autoref{tab_logs}. These logs represent diverse organizational processes from various domains, including financial services, healthcare, education, and manufacturing, all of which are typically used in PPM studies.

\begin{table}[t]
\centering
\scriptsize
\caption{Characteristics of the employed event logs: ACT = distinct activities, RES = distinct resources, ALT = average lead time of process instances (days).}
\begin{tabular}{lrrrrrr}
\toprule
Event log & \multicolumn{1}{c}{Cases} & \multicolumn{1}{c}{Events} & \multicolumn{1}{c}{ACT} & \multicolumn{1}{c}{RES} & \multicolumn{1}{c}{Variants} & \multicolumn{1}{c}{ALT} \\
\midrule
P2P & 608 & 9119 & 21 & 27 & 70 & 21.5 \\
Production & 225 & 4503 & 24 & 41 & 217 & 20.6 \\
Helpdesk & 4580 & 21348 & 14 & 22 & 226 & 40.9 \\
Sepsis & 1050 & 15214 & 16 & 26 & 846 & 28.5 \\
BPIC20DD & 10500 & 56437 & 17 & 7 & 99 & 11.5 \\
BPIC20ID & 6449 & 72151 & 34 & 8 & 753 & 86.5 \\
BPIC12C & 13087 & 164506 & 23 & 69 & 4336 & 8.6 \\
BPIC15-1 & 1199 & 52217 & 398 & 23 & 1170 & 95.9 \\
BPIC13I & 7554 & 65533 & 13 & 1440 & 2278 & 12.1 \\
\bottomrule
\end{tabular}
\label{tab_logs}
\end{table}

\mypar{Data Split}
Each event log is split temporally into 64\% train, 16\% validation, and 20\% test set. All models are trained on event prefixes of each case, ranging from length 1 to $n-1$, where $n$ denotes the total number of events in the case.
Additionally, we evaluate the robustness of our findings using the unbiased splitting strategy proposed by \textcite{WeytjensW21a}, which introduces stricter safeguards against data leakage. The corresponding results, provided in our repository, confirm that the main observations of this paper also hold under this splitting strategy.

\mypar{Feature Extraction}
For each prefix, we construct a fixed-length sequence representation over the maximum case length. Each event is represented by activity, resource, weekday, hour of day, time since the previous event, and elapsed time since case start. Categorical attributes are label-encoded, while time-based features are converted to days, log-transformed, and standardized. Shorter prefixes are left-padded with zeros.

\mypar{PPM Tasks}
We evaluate three common PPM tasks: next activity prediction (NAP), next timestamp prediction (NTP), and remaining time prediction (RTP). Our evaluation includes all combinations of these three tasks, i.e., three STL models and four MTL models.

\mypar{Architectures}
Deep learning models for PPM predominantly rely on LSTMs, Transformers, and CNNs. We therefore base our evaluation on these three architectures. Each model consists of a shared encoder (2–3 layers) and task-specific linear heads of size 128. The LSTM encoder comprises two layers with 128 units and one dropout layer between them. The CNN encoder includes three convolutional layers (64 filters, kernel size 3), each followed by max pooling and dropout. The Transformer encoder consists of two multi-head attention layers with 8 heads. A dropout rate of 0.1 is applied throughout.

\mypar{Training Setup}
All models are trained using cross-entropy loss for NAP and L1 loss for both NTP and RTP. Training is performed using the Adam optimizer for up to 200 epochs, with early stopping patience of 10 epochs.

\mypar{MTO Methods}
To balance the three tasks, we evaluate 13 different MTO methods, all of which are introduced in~\autoref{sec:background}. This selection covers a broad range of both loss-weighting and gradient-based approaches.

\mypar{Hyperparameters}
Following recent insights from the MTL literature \parencite{xin2022current,KirchdorferEKSSK22}, we acknowledge that different MTO methods require distinct learning rates to reach their specific optimum. Thus, we conduct a grid search over various combinations of learning rates, specifically $\lambda \in \{0.0001, 0.001, 0.01\}$, along with relevant MTO-specific hyperparameters (see details in our repository). We select the best model based on validation loss. We also tune $\lambda$ for the STL models.


\mypar{Evaluation Metrics}
We use both task-specific metrics---\emph{accuracy} for NAP and \emph{MAE} for NTP/RTP---and the aggregate $\Delta_m$-metric, which measures the average relative performance gain of the multi-task model $M_m$ w.r.t.\ a single-task baseline $M_b$:
    $\Delta_m = \frac{1}{K}\sum_{k=1}^K(-1)^{l_k}(M_{m,k}-M_{b,k})/M_{b,k} * 100$.
Here, $l_k$ is $1 / 0 $ if a higher / lower value is better for criterion $M_k$, i.e., 1 for NAP and 0 for NTP and RTP. A negative $\Delta_m$ score indicates that, on average for all tasks, MTL outperforms STL.  




\section{Results and Discussion}
\label{sec:results}
This section presents the experimental results evaluating the effectiveness and potential of MTL in the context of PPM. Guided by the four research questions introduced in~\autoref{sec:intro}, we perform dedicated experiments and analyses for each and discuss the corresponding findings.

\subsection{RQ1: MTL vs. STL}
\label{sec:rq1}
\begin{figure} 
    \centering
    \includegraphics[width=\linewidth]{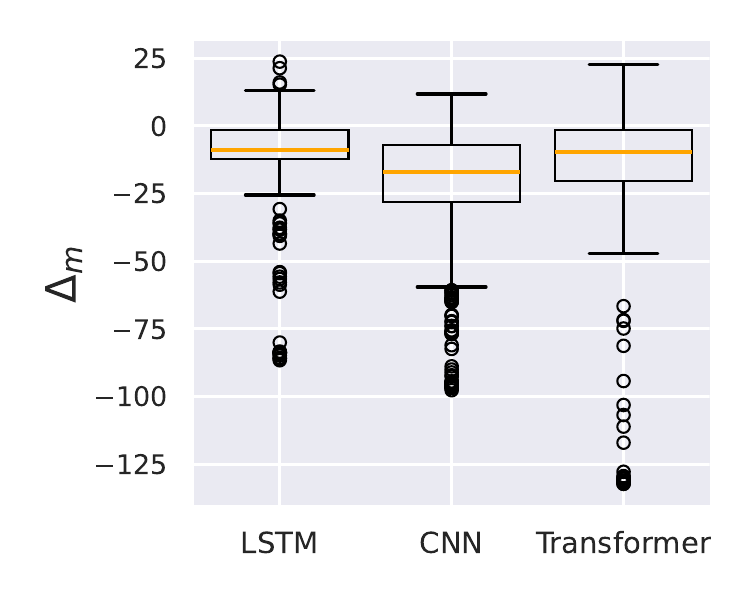}
    \caption{MTL performance by model architecture, measured by $\Delta_m$.}
    \label{fig:1}
\end{figure}

To address RQ1, we examine whether a multi-task model can outperform three separate single-task models across the three PPM tasks. We conduct this analysis on all nine datasets and report the results per architecture in \autoref{fig:1}. The figure presents box plots of the resulting $\Delta_m$ scores, illustrating their distribution across all datasets, MTO methods, and multi-task combinations. Concretely, for each architecture (e.g., the LSTM), we evaluate nine datasets; for each dataset, we train four multi-task configurations (NAP+NTP+RTP, NAP+NTP, NAP+RTP, NTP+RTP), and each multi-task configuration is trained with 13 different MTO methods, each with three random seeds. In total, this results in 1,404 multi-task models per architecture.

Across all three architectures, both the first and third quartiles lie below 0, indicating that MTL outperforms the corresponding single-task models in the majority of experiments. However, every architecture also shows instances where $\Delta_m$ exceeds 0, meaning MTL can underperform its single-task counterparts in certain cases. The following sections will provide more detailed insights, examining whether these cases stem from specific task combinations, particular datasets, or poorly performing MTO methods.
Yet, overall, the results highlight the generally positive performance impact of multi-task training for PPM compared to STL.

Although \autoref{fig:1} shows minor distributional differences across the three architectures and a Wilcoxon signed-rank test indicates a statistically significant difference between the LSTM and CNN ($p = 0.0391$), the benefits of MTL remain largely consistent regardless of the architecture.


Note that, given the large number of experiments, the following analyses often include only representative subsets. The complete set of results is provided in the supplementary material available in our repository.

\subsection{RQ2: Impact of MTL on Individual Tasks}
\label{sec:rq2}

Building on the findings from \autoref{sec:rq1}, which showed that MTL generally outperforms STL on average across the three PPM tasks—next activity prediction (NAP), next time prediction (NTP), and remaining time prediction (RTP)—we now address RQ2 by examining how MTL affects each task individually.

\mypar{Statistical Comparison}
We test whether MTL yields statistically significant performance differences from STL for each task in \autoref{tab:wilcoxon}, reporting p-values of the Wilcoxon signed-rank test.

\begin{table}[t!]
    \centering
    \caption{Comparison of MTL and STL using the Wilcoxon signed-rank test. P-values are based on all task combinations and MTO methods.}
    \setlength{\tabcolsep}{6pt}
    \begin{tabular}{l r r r}
        \toprule
        Model & NAP & NTP  & RTP\\
        \midrule
        LSTM  & \textbf{0.004} & 0.250 & 0.055\\
        CNN  & \textbf{0.004} & 0.570 & 0.734\\
        Transformer & \textbf{0.008}  & \textbf{0.008} & 0.742\\
        \bottomrule    
    \end{tabular}
    \label{tab:wilcoxon}
\end{table}

The results show that—regardless of the underlying architecture—MTL leads to a statistically significant difference at the 99\% confidence level for the NAP task. In contrast, for both NTP and RTP, the statistical tests indicate no significant differences between MTL and STL in most cases, with the sole exception of NTP under the Transformer architecture. Overall, these findings suggest that joint learning neither consistently benefits nor harms the two temporal tasks.
A possible explanation is that the temporal tasks derive limited additional benefit from the shared representation beyond their own supervision signals. Meanwhile, the absence of performance degradation indicates limited task interference, suggesting the shared architectures have sufficient capacity for joint learning without compromising temporal predictions. We discuss this interpretation further in \autoref{sec:rq4}.

Taken together, since MTL improves the average performance over STL and NAP is the only task exhibiting consistent and significant differences, \emph{the overall performance improvements of MTL appear to be primarily driven by accuracy enhancements in the NAP task}. We further analyze this effect next.

\mypar{Detailed LSTM Results}
To make the impact of MTL more concrete, \autoref{tab:results_main} reports the performance of the LSTM models—trained jointly on all three tasks—for each of the nine datasets. Alongside the STL baseline, we present results for the EW method as well as the best- and worst-performing MTO approaches. Across nearly all configurations, the $\Delta_m$ values are negative, confirming that MTL improves the average task performance relative to STL.

\begin{table}[t!]
    \centering
    \caption{Results for LSTM. For each log, we show the average results over three random seeds (and the standard deviation) for STL, EW, and the best/worst MTO method according to $\Delta_m$. 
    MAE reported in days.
    }
    \setlength{\tabcolsep}{2.5pt}
    \scriptsize
    \renewcommand{\arraystretch}{1.2}
    \begin{tabular}{ccrrrrr}
        \toprule
        \multirow{2}{*}{Log} & \multirow{2}{*}{MTO} & \multicolumn{2}{c}{NAP} & NTP & RTP & \multirow{2}{*}{$\Delta_m$ $\downarrow$} \\
         & & ACC $\uparrow$ & macro-F1 $\uparrow$& MAE $\downarrow$ & MAE $\downarrow$ & \\

        \midrule
        \multirow{4}{*}{\rotatebox{90}{P2P}} & STL  & 0.73 (0.01) & 0.65 (0.01) & 1.07 (0.08)  & 4.47 (0.11) & / \\
        \noalign{\vskip 1mm}
        \cdashline{2-7}
        \noalign{\vskip 1mm}
         & EW  & 0.86 (0.00) & 0.78 (0.01) & 1.08 (0.07)  & 4.73 (0.15) & -4.19 \\
         & GradDrop  & 0.87 (0.01) & 0.78 (0.01) & 1.03 (0.02)  & 4.41 (0.09) & -8.05 \\
         & IMTL  & 0.85 (0.01) & 0.78 (0.01) & 1.07 (0.08)  & 5.13 (0.52) & -1.09 \\

        \midrule
        \multirow{4}{*}{\rotatebox{90}{Production}} & STL  & 0.44 (0.01) & 0.21 (0.01) & 0.99 (0.16)  & 4.72 (0.29) & / \\
        \noalign{\vskip 1mm}
        \cdashline{2-7}
        \noalign{\vskip 1mm}
         & EW  & 0.49 (0.02) & 0.26 (0.02) & 1.01 (0.15)  & 4.77 (0.15) & -2.50 \\
         & GradDrop  & 0.50 (0.02) & 0.27 (0.01) & 1.00 (0.15)  & 4.79 (0.03) & -3.44 \\
         & IMTL  & 0.50 (0.02) & 0.29 (0.02) & 1.08 (0.14)  & 6.69 (2.10) & 12.89 \\

        \midrule
        \multirow{4}{*}{\rotatebox{90}{Helpdesk}} & STL  & 0.68 (0.01) & 0.25 (0.02) & 5.09 (0.10)  & 6.38 (0.06) & / \\
        \noalign{\vskip 1mm}
        \cdashline{2-7}
        \noalign{\vskip 1mm}
         & EW  & 0.80 (0.00) & 0.33 (0.02) & 5.07 (0.15)  & 6.33 (0.10) & -6.43 \\
         & GradDrop  & 0.80 (0.01) & 0.30 (0.02) & 5.06 (0.10)  & 6.33 (0.06) & -6.47 \\
         & IMTL  & 0.80 (0.01) & 0.32 (0.03) & 5.10 (0.09)  & 6.35 (0.05) & -5.98 \\

        \midrule
        \multirow{4}{*}{\rotatebox{90}{Sepsis}} & STL  & 0.48 (0.01) & 0.38 (0.01) & 1.55 (0.02)  & 19.28 (1.42) & / \\
        \noalign{\vskip 1mm}
        \cdashline{2-7}
        \noalign{\vskip 1mm}
         & EW  & 0.58 (0.01) & 0.45 (0.01) & 1.18 (0.14)  & 18.58 (1.35) & -16.09 \\
         & NashMTL  & 0.58 (0.01) & 0.43 (0.01) & 1.10 (0.10)  & 18.66 (1.80) & -17.92 \\
         & UW  & 0.60 (0.01) & 0.46 (0.01) & 1.62 (0.04)  & 23.21 (0.78) & -0.55 \\

        \midrule
        \multirow{4}{*}{\rotatebox{90}{BPI20DD}} & STL  & 0.71 (0.01) & 0.33 (0.02) & 1.64 (0.03)  & 2.80 (0.02) & / \\
        \noalign{\vskip 1mm}
        \cdashline{2-7}
        \noalign{\vskip 1mm}
         & EW  & 0.86 (0.01) & 0.42 (0.02) & 1.64 (0.03)  & 2.79 (0.03) & -7.20 \\
         & DWA  & 0.86 (0.00) & 0.42 (0.02) & 1.64 (0.02)  & 2.78 (0.03) & -7.25 \\
         & GLS  & 0.86 (0.01) & 0.42 (0.02) & 1.65 (0.02)  & 2.82 (0.02) & -6.58 \\

        \midrule
        \multirow{4}{*}{\rotatebox{90}{BPI20ID}} & STL  & 0.62 (0.00) & 0.25 (0.00) & 5.72 (0.04)  & 15.48 (0.30) & / \\
        \noalign{\vskip 1mm}
        \cdashline{2-7}
        \noalign{\vskip 1mm}
         & EW  & 0.85 (0.00) & 0.44 (0.01) & 5.60 (0.05)  & 15.54 (0.20) & -12.90 \\
         & UW-SO  & 0.85 (0.00) & 0.47 (0.01) & 5.54 (0.05)  & 15.40 (0.10) & -13.65 \\
         & UW-O  & 0.86 (0.00) & 0.46 (0.02) & 5.73 (0.05)  & 16.06 (0.19) & -11.51 \\

        \midrule
        \multirow{4}{*}{\rotatebox{90}{BPIC12C}} & STL  & 0.65 (0.01) & 0.37 (0.01) & 0.58 (0.01)  & 4.66 (0.12) & / \\
        \noalign{\vskip 1mm}
        \cdashline{2-7}
        \noalign{\vskip 1mm}
         & EW  & 0.79 (0.00) & 0.64 (0.00) & 0.59 (0.02)  & 5.21 (0.02) & -2.04 \\
         & UW-SO  & 0.79 (0.00) & 0.64 (0.01) & 0.59 (0.02)  & 5.21 (0.00) & -2.04 \\
         & UW  & 0.78 (0.00) & 0.64 (0.00) & 0.60 (0.01)  & 5.84 (0.03) & 2.89 \\

         \midrule
        \multirow{4}{*}{\rotatebox{90}{BPIC15-1}} & STL  & 0.17 (0.01) & 0.04 (0.00) & 2.03 (0.20)  & 13.65 (0.44) & / \\
        \noalign{\vskip 1mm}
        \cdashline{2-7}
        \noalign{\vskip 1mm}
         & EW   & 0.30 (0.01) & 0.12 (0.00) & 2.09 (0.22)  & 14.41 (0.49) & -23.09 \\
         & UW-O   & 0.41 (0.01) & 0.21 (0.00) & 1.92 (0.01)  & 16.47 (1.49) & -43.49 \\
         & UW-SO   & 0.24 (0.00) & 0.08 (0.00) &  2.10 (0.23) & 18.02 (1.00) & -3.67 \\

        \midrule
        \multirow{4}{*}{\rotatebox{90}{BPIC13I}} & STL  & 0.52 (0.00) & 0.18 (0.01) & 1.40 (0.06)  & 9.65 (0.30) & / \\
        \noalign{\vskip 1mm}
        \cdashline{2-7}
        \noalign{\vskip 1mm}
         & EW  & 0.64 (0.01) & 0.27 (0.01) & 1.34 (0.06)  & 9.32 (0.31) & -10.69 \\
         & GradNorm  & 0.66 (0.00) & 0.31 (0.02) & 1.32 (0.07)  & 9.41 (0.17) & -12.25 \\
         & IMTL  & 0.56 (0.06) & 0.19 (0.01) & 1.54 (0.05)  & 12.17 (0.26) & 8.99 \\

        \bottomrule    
    \end{tabular}
    \label{tab:results_main}
\end{table}

A closer inspection of the individual task metrics, however, shows that these gains are predominantly driven by substantial improvements in NAP. In most datasets, NAP accuracy increases by roughly 10 percentage points—for example, in the \textit{Sepsis} dataset, accuracy rises from 48\% to 60\%. While some improvements are more modest, such as 6 percentage points in the \textit{Production} dataset, others are considerably larger, including an increase of 24 percentage points in \textit{BPI20ID}.
In contrast, the performance differences for NTP and RTP between STL and MTL are generally small and often lie within one standard deviation. For instance, in \textit{Sepsis}, MTL with EW reduces the NTP error from 1.55 days to 1.18 days and the RTP error from 19.28 days to 18.58 days. However, other MTO methods—such as UW here—can lead to substantial degradations, as seen in the same dataset where UW increases the RTP error to 23.21 days.

\begin{figure}[t!]
    \centering
    \includegraphics[width=\linewidth]{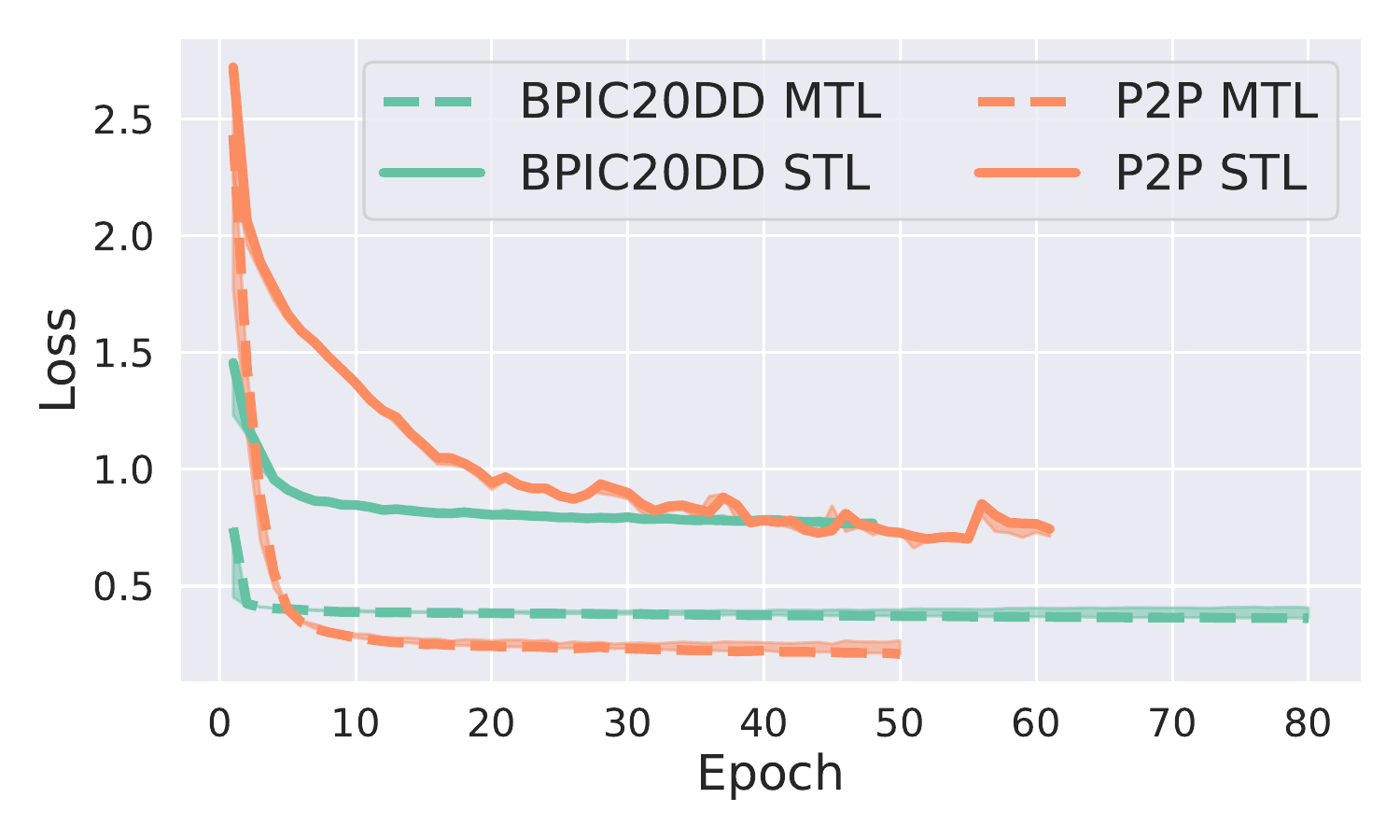}
    \caption{Comparison of NAP's train and validation loss for STL and MTL (with EW) on two datasets. The lines represent the train loss and the small shaded area captures the gap to the validation loss, respectively.}
    \label{fig:loss_comparison}
\end{figure}

\mypar{Analysis of Overfitting}
Since MTL is often claimed to enhance generalization performance \parencite{caruana1997multitask}, we examine whether the weaker single-task results stem from overfitting. As illustrated for two example datasets in \autoref{fig:loss_comparison}, both the training and validation losses of the NAP task are substantially higher for the STL model than for the MTL model (based on the LSTM).
This indicates that the inferior STL performance is not due to overfitting; rather, both losses suggest that STL struggles to learn equally informative representations. MTL’s superior performance is therefore more consistent with an inductive transfer effect: auxiliary temporal tasks provide additional supervision on process progression and execution dynamics, enabling the shared encoder to learn representations better suited for distinguishing the next activity. As a result, MTL converges faster and reaches a better solution than STL.

\begin{figure}[t!]
    \centering
    \includegraphics[width=\linewidth]{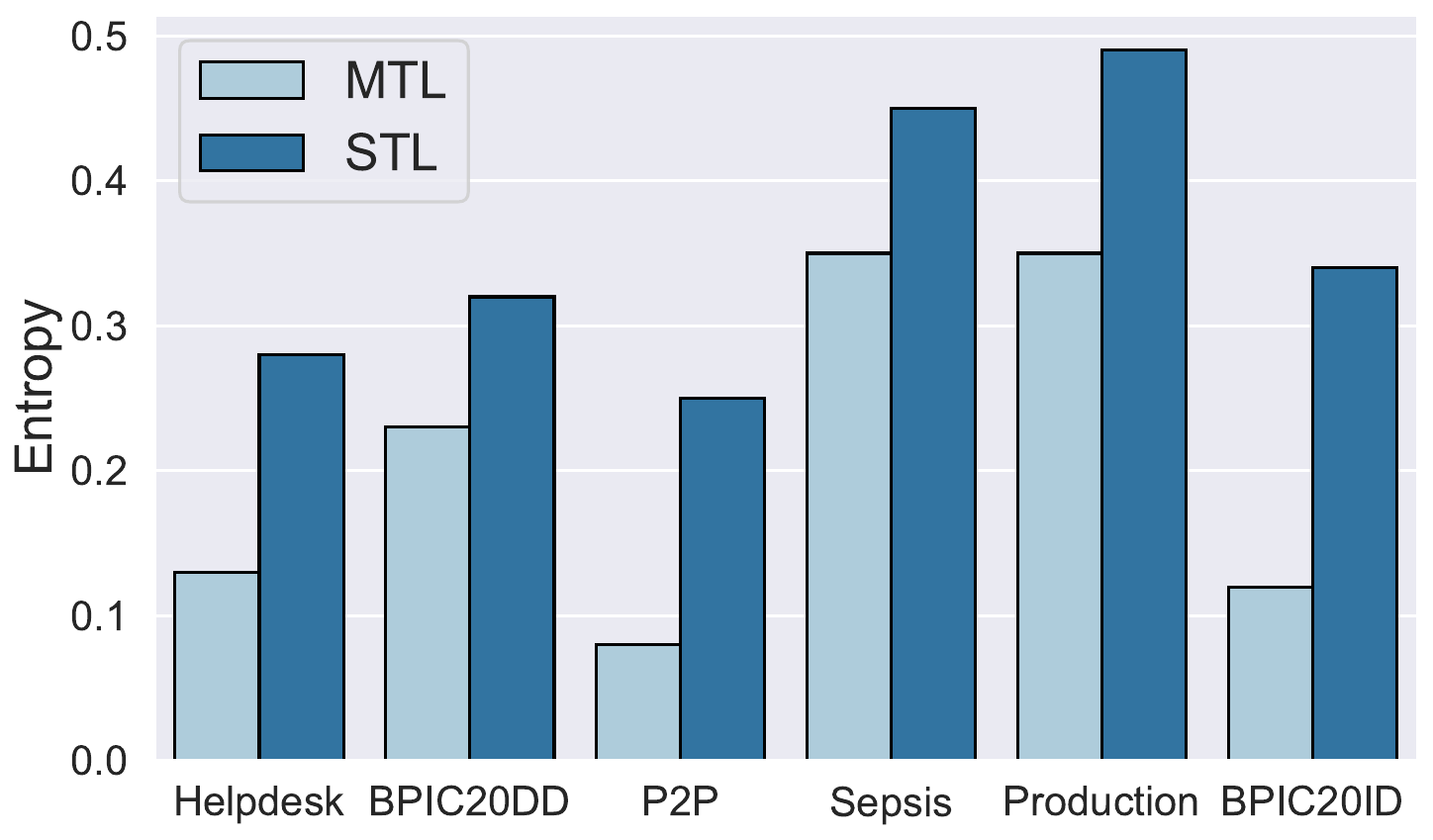}
    \caption{Average model uncertainty per dataset on the respective test set, measured by the normalized entropy. Results are based on LSTM models with EW.}
    \label{fig:entropy}
\end{figure}

\mypar{Impact on Predictive Certainty}
We further examine the model’s predictive certainty by comparing the normalized entropy 
of the NAP predictions between STL and MTL models across datasets in~\autoref{fig:entropy}. MTL consistently yields lower entropy, indicating the model's higher confidence. This reduced uncertainty is a strong signal of better generalization. In multi-class classification problems like NAP, high entropy often reflects ambiguity or poor internal representations \parencite{postels2020hidden}.
These findings provide a plausible explanation for the superior NAP performance of MTL. By jointly learning temporal prediction tasks, the shared encoder receives additional supervision about process progression, reducing ambiguity between competing activity classes and leading to lower predictive entropy. While NTP and RTP themselves benefit little from joint training, they appear to strengthen the learning signal for NAP by providing complementary temporal context.

\mypar{Impact on Infrequent Class Performance}
As discussed in \autoref{sec:intro}, class imbalance poses a major challenge for NAP \parencite{NeuLF22}: existing PPM models often overfit to frequent activities—achieving high overall accuracy—while performing poorly on rare yet operationally critical ones. Consequently, several methods have been proposed or leveraged to address this problem in the single-task setting, including cost-sensitive learning \parencite{nguyen2020time}, class-balanced focal loss \parencite{He2025}, and class undersampling \parencite{GrohsPR23}. Given that MTL improves overall NAP accuracy, we now investigate whether these gains also extend to infrequent classes.

To this end, we compare the macro-F1 scores achieved by STL and MTL models in \autoref{tab:results_main}. Because macro-F1 assigns equal weight to each class, it is particularly sensitive to performance on rare activities and therefore well-suited for assessing this effect. Notably, the MTL models outperform their STL counterparts across all datasets on this metric, often by margins of 10–15 points. These results indicate that MTL inherently mitigates class imbalance, without requiring specialized loss functions or additional sampling strategies. 

\begin{figure}[t!]
    \centering
    \includegraphics[width=\linewidth]{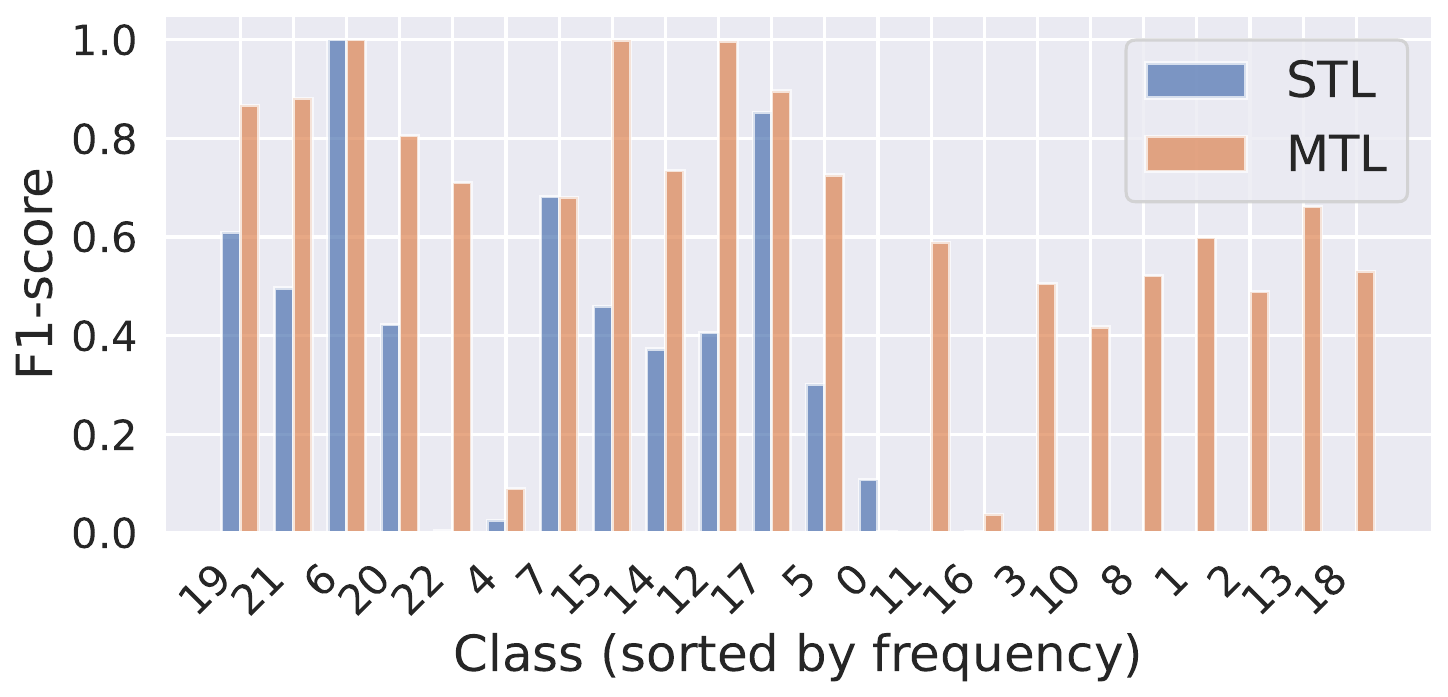}
    \caption{Per-class F1-score comparison between MTL and STL models using CNN architecture for BPIC12C. The MTL model employs the NAP+NTP+RTP task combination. Activity classes are sorted by frequency.}
    \label{fig:class_analysis_12C}
\end{figure}

To make this more tangible,~\autoref{fig:class_analysis_12C} compares per-class F1-scores of MTL and STL for the loan application log BPIC12C (using CNN). MTL consistently outperforms STL, not only for frequent activities but also for infrequent ones. While STL struggles on rare activities due to limited training data, MTL leverages auxiliary signals from time-related tasks, providing contextual information that enhances generalization to such cases.
For example, the infrequent class 1 (receiving an offer response from the applicant), is never predicted correctly by STL, with MTL providing more reliable estimates. 
A t-test further indicates that this activity is associated with significantly shorter next timestamps. This finding supports the hypothesis that temporal tasks provide complementary information that helps distinguish otherwise difficult activity classes, particularly those with limited training examples.

\mypar{Prefix-Length Analysis}
To assess whether MTL effects are consistent throughout process execution, we analyzed performance as a function of prefix length (see supplementary material). The gains are not concentrated in a narrow prefix length range. For NAP, MTL consistently outperforms STL across virtually all prefix lengths and datasets, with the advantage emerging early and remaining stable as longer prefixes are incorporated. Effects on NTP and RTP are more heterogeneous: improvements are often modest and depend more strongly on the event log, architecture, and MTO method. Overall, this confirms that our main conclusions are robust across different stages of process execution rather than driven by specific prefix lengths.

\mypar{Implications for Additional PPM Tasks}
Our findings may have implications for other PPM tasks like suffix and outcome prediction.
Suffix prediction is commonly performed autoregressively via repeated next-activity (and timestamp) predictions until case completion. Given MTL’s substantial improvements for NAP and its neutral effect on temporal predictions, MTL could also benefit suffix prediction.
Our findings may also transfer to outcome prediction: when outcomes are defined by the occurrence of specific future activities, MTL is likely beneficial, given its strong gains for NAP; when outcomes depend primarily on temporal quantities, such as the time between process milestones, we expect limited benefits, consistent with the small gains observed for temporal tasks. Testing these hypotheses is an interesting direction for future work.

\subsection{RQ3: Impact of Different Task Combinations} 
\label{sec:rq3}
\begin{figure}[t!] 
    \centering
    \includegraphics[width=\linewidth]{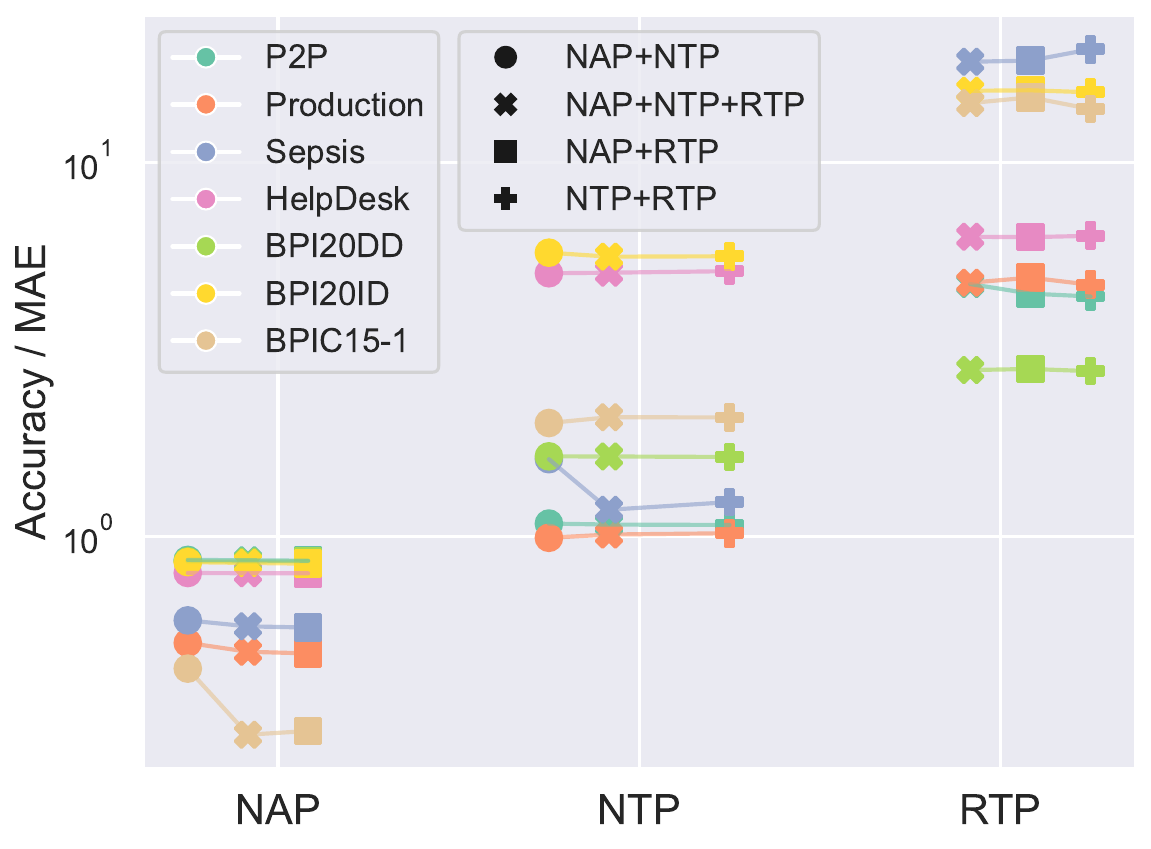}
    \caption{Impact of task combinations on performance. Results are based on EW with the LSTM model.}
    \label{fig:task_combinations}
\end{figure}

An important line of research in MTL focuses on task affinities---determining which tasks benefit from joint learning through shared representations. This is particularly relevant given the risk of negative transfer, where training multiple tasks simultaneously can impair performance.
We therefore formulate RQ3, investigating whether different task combinations lead to significant performance variations. \autoref{fig:task_combinations} presents the test metrics
for each of the three tasks across all four possible multi-task combinations, using the LSTM architecture with EW as a representative example.
Overall, for the majority of datasets, we observe little to no performance variation between task combinations. The few exceptions---such as a minor drop in NAP accuracy on BPIC15-1---are rare, marginal, and lack a consistent pattern across datasets. This contrasts with \textcite{Camargo2019} who reported that combining categorical and temporal PPM tasks tends to degrade performance. Our experiments—across CNN, LSTM, and Transformer architectures (see supplementary material)—instead show that the choice of task combination has minimal impact.  This suggests that the benefits of MTL do not arise from a particular task pairing but from the presence of complementary supervisory signals more generally.


\subsection{RQ4: Comparison of MTO Methods} 
\label{sec:rq4}
A wide range of MTO methods has been proposed to enhance the training of MTL models; however, mostly in the context of computer vision. Therefore, we address RQ4 by comparing the performance of several widely used MTO methods, investigating which---if any---are specifically suitable for PPM applications.

\begin{figure}[t!]
    \centering
    \includegraphics[width=\linewidth]{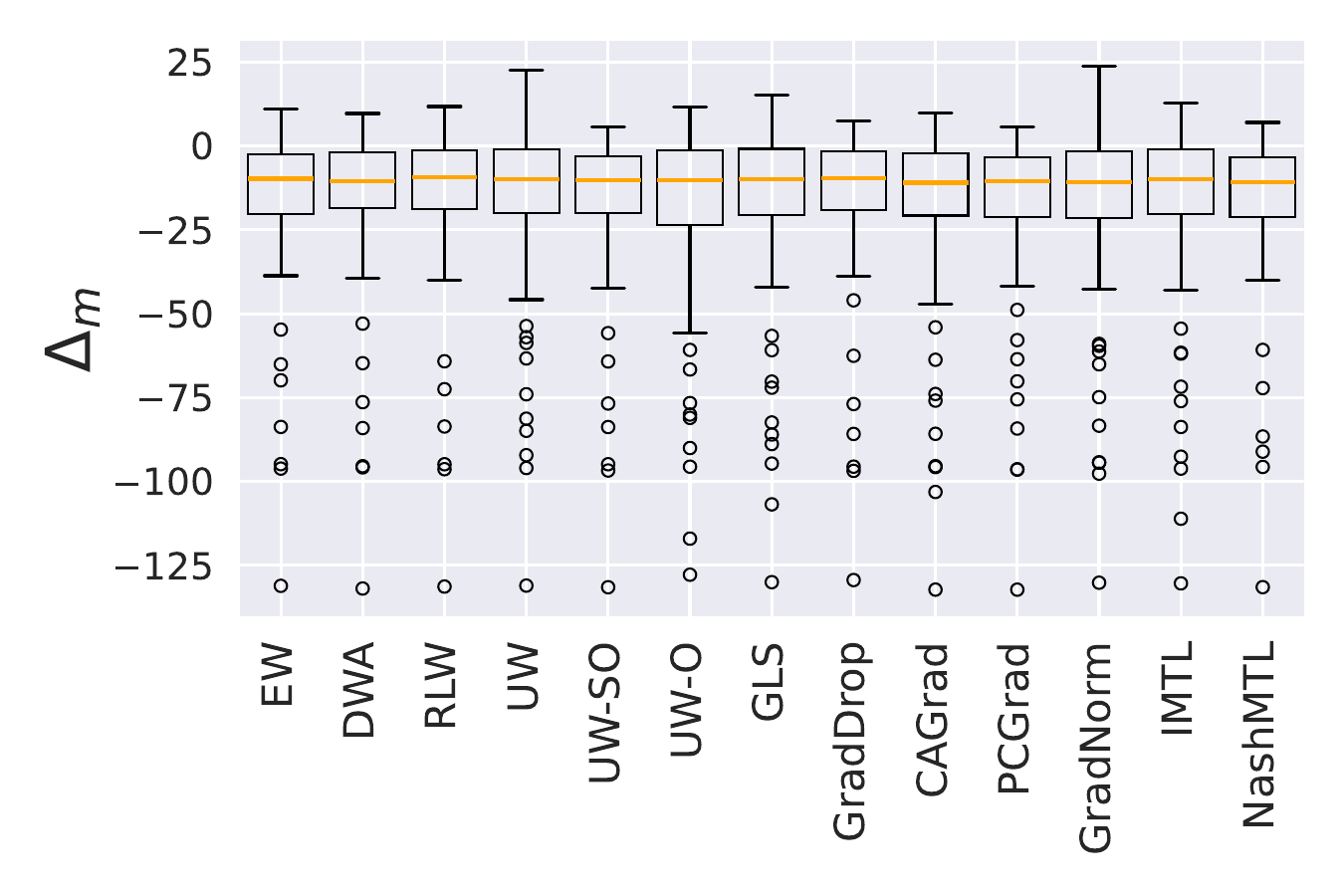}
    \caption{Box plots showing $\Delta_m$ per MTO method, aggregated over all architectures and task combinations.}
    \label{fig:mto}
\end{figure}

\mypar{Statistical Comparison}
\autoref{fig:mto} presents box plots showing the distribution of $\Delta_m$ scores for each MTO method, aggregated across all datasets, model architectures, and task combinations. Interestingly, all thirteen methods exhibit very similar performance, with median $\Delta_m$ values clustering around -10 and interquartile ranges generally spanning from 0 to -20. 
While some methods, such as UW and GradNorm, display slightly wider spreads, the overall differences between methods are generally modest. 
A statistical test confirms this lack of significant differences between methods (see Appendix).
These results suggest that even simple methods, such as equal weighting (EW), perform comparably to more sophisticated ones.
Note that we also benchmarked the computationally expensive Scalarization method on several smaller datasets, including P2P. However, despite its high cost, it did not outperform the other methods. Consequently, we excluded it from further analysis.
We hypothesize that the limited impact of the MTO choice stems from (a) the low degree of task interference, and (b) the use of sufficiently large models, which provide enough capacity such that the tasks do not need to compete for shared resources. We investigate both hypotheses next.




\mypartwo{Hypothesis (a)}
To empirically examine the first hypothesis, we measure the cosine similarities between task gradients in the shared backbone (encoder) during training. A negative cosine similarity (i.e., an angle greater than $90^\circ$) indicates potential negative transfer due to task interference, particularly when one gradient has a significantly larger magnitude than the other \parencite{YuK0LHF20}. 
Aggregating cosine values across datasets and epochs, we observe values slightly above $90^\circ$, i.e., near-orthogonal to mildly conflicting interactions.
For instance, in the fully shared CNN model with EW, the average angles between task gradients are $92^\circ$ (NAP–NTP), $100^\circ$ (NTP–RTP), and $106^\circ$ (NAP–RTP). 
These measurements suggest that task gradients are close to orthogonal but with some conflict, which may limit (but not preclude) negative transfer.

\begin{figure}[]
    \centering
    \includegraphics[width=\linewidth]{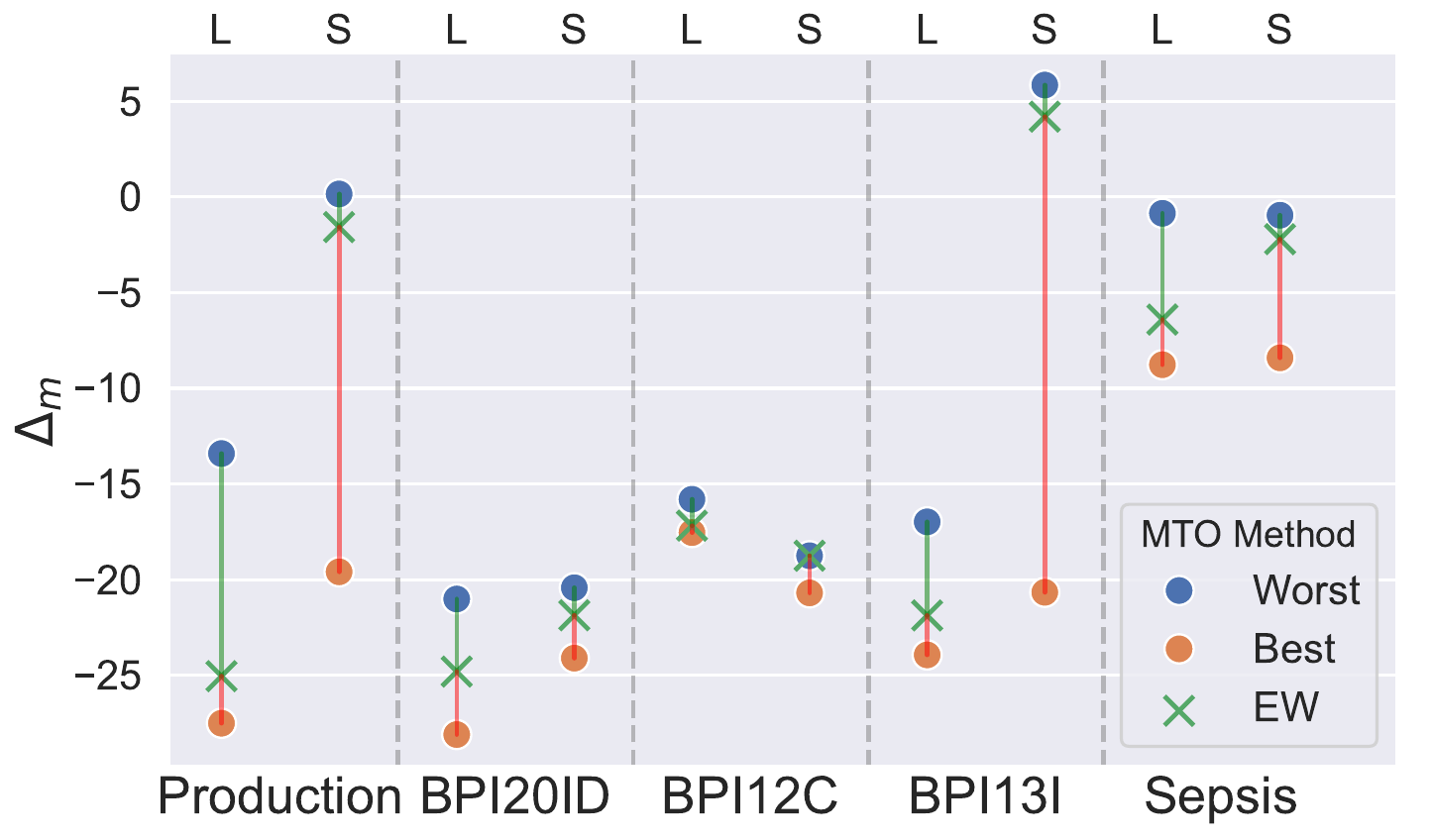}
    \caption{$\Delta_m$ performance for the large (L) and small (S) variants of the CNN model. We show the best and worst performing MTO methods as well as EW.}
    \label{fig:small_cnn}
\end{figure}

\mypartwo{Hypothesis (b)}
To further investigate the second hypothesis, we conduct an ablation study comparing the performance of MTO methods on our original CNN model and a smaller variant. Specifically, we reduce the encoder depth from three convolutional and pooling layers to just one. \autoref{fig:small_cnn} shows the distribution of $\Delta_m$ values for each MTO method on both the large (L) and small (S) CNN models.
A consistent pattern emerges across most datasets: for the larger model, EW performs comparably to the best MTO methods, while for the smaller model, it ranks among the worst. This discrepancy is less critical for simpler datasets such as BPI12C, where the performance gap between MTO methods is small. However, for more challenging datasets like Production or BPI13I, the underperformance of EW becomes significant, with differences reaching up to 25 $\Delta_m$ points. An analysis of MTO performance in the small model variant reveals that simple loss weighting methods such as EW, DWA, and RLW generally underperform, while approaches like inverse loss weighting (UW-O) and the gradient-based NashMTL consistently achieve strong results.
E.g., on BPI13I, EW attains a $\Delta_m$ of 4.2 compared to -20.7 for UW-O due to worse performance on all three tasks (NAP: 0.39 vs. 0.45; NTP: 2.57 vs. 1.58; RTP: 15.93 vs. 12.30; details in repository).

In summary, the findings suggest that model capacity affects task interference, i.e., the importance of task balancing and the choice of MTO method is minimal when models are sufficiently large and the data is easy to learn. However, in resource-constrained settings, effective task balancing becomes essential, especially for complex datasets. Given that gradient-based methods tend to incur higher training costs, 
our results advocate for inverse loss weighting (UW-O) \parencite{KirchdorferEKSSK22} as a practical compromise: it is simple, computationally efficient, and performs reliably in low-capacity scenarios.


\section{Conclusion}
\label{sec:conclusion}
This paper presents the first systematic evaluation of multi-task learning (MTL) in predictive process monitoring (PPM), addressing questions about its effectiveness and practical value. 
Through extensive experiments, we find that MTL yields substantial improvements compared to its single-task counterpart, mostly enhancing the next-activity prediction. In contrast, temporal tasks (next and remaining time prediction) are hardly affected. 
Our analysis suggests that these gains stem from inductive transfer. In particular, auxiliary temporal tasks provide complementary supervision, enabling the shared encoder to learn richer representations. This effect is strongest for next-activity prediction, which suffers from class imbalance.

\mypar{Implications}
Table~\ref{tab:findings_recommendations} summarizes our key findings and recommendations for practitioners.
Overall, our findings establish MTL as a competitive and cost-efficient alternative to the single-task setup, with implications for both academia and industry. For process mining vendors, MTL can reduce the number of predictive models that need to be trained and maintained across customers, processes, and prediction tasks. For practitioners, our results suggest adopting MTL, particularly when next-activity prediction is of primary interest. For researchers, our findings open several avenues for future work, including investigating the potential of MTL in other process mining applications such as simulation or anomaly detection, and developing PPM-specific MTL approaches.



\begin{table}[t]
\centering
\scriptsize
\renewcommand{\arraystretch}{1.2}
\setlength{\tabcolsep}{2.5pt}
\caption{Key findings and practical recommendations.}
\label{tab:findings_recommendations}
\begin{tabular}{p{0.43\linewidth}p{0.50\linewidth}}
\toprule
\textbf{Finding} & \textbf{Recommendation} \\
\midrule
MTL mainly improves NAP, only minimal impact on temporal tasks. & Use MTL particularly when activity prediction is central. \\
MTL improves performance on infrequent activities.
& Consider MTL when rare but relevant activities matter. \\
Equal weighting is often competitive.
& Start with simple EW before using more complex MTO methods. \\
Task balancing matters under limited model capacity.
& Use adaptive weighting when models are small or data is complex. \\
Task-wise effects can differ.
& Monitor each task separately rather than relying only on aggregate scores. \\
\bottomrule
\end{tabular}
\end{table}

\printbibliography

\end{document}